\documentclass{article}

\usepackage{main/spconf}

\usepackage{xspace}
\usepackage{algorithmic}
\usepackage{array}
\usepackage{textcomp}
\usepackage{stfloats}
\usepackage{url}
\usepackage{graphicx}
\usepackage{amsmath,amssymb,amsfonts}
\usepackage{multirow,multicol}
\usepackage{todonotes}
\usepackage{microtype}
\usepackage{overpic}
\usepackage{booktabs}
\usepackage{enumitem}
\usepackage{textcomp}
\usepackage[table,xcdraw]{xcolor}
\usepackage{pgfplots}
\pgfplotsset{compat=1.18}
\usepackage{pifont}
\usepackage{wrapfig}

\title{Generative 6D pose estimation via conditional flow matching}

\name{
Amir Hamza$^{1,2}$ \quad
Davide Boscaini$^{1}$ \quad
Weihang Li$^{3,4}$ \quad
Benjamin Busam$^{3,4}$ \quad
Fabio Poiesi$^{1}$\vspace{-3mm}
}
\address{
$^{1}$Fondazione Bruno Kessler \quad
$^{2}$University of Trento \quad
$^{3}$Technical University of Munich \\
$^{4}$Munich Center for Machine Learning \vspace{-3mm}
}

\begin{document}
\newcommand{\fabio}[1]{\todo[color=red!20, inline, author=Fabio]{#1}}
\newcommand{\davide}[1]{\todo[color=yellow!20, inline, author=Davide]{#1}}
\newcommand{\andrea}[1]{\todo[color=blue!20, inline, author=Andrea]{#1}}

\newcommand{\amir}[1]{\todo[color=green!20, inline, author=Amir]{#1}}

\newcommand{\warning}[1]{\textbf{\color{red!90}{#1}}}

\newcommand{\acronym}{Flose\xspace}

\definecolor{tableazure}{RGB}{214,234,248}
\definecolor{forestgreen}{RGB}{34,139,34}

\newcommand{\cmark}{\ding{51}}
\newcommand{\xmark}{\ding{55}}


\maketitle

\vspace{-1.5mm}
\begin{abstract}
Existing methods for instance-level 6D pose estimation typically rely on neural networks that either directly regress the pose in $\mathrm{SE}(3)$ or estimate it indirectly via local feature matching.
The former struggle with object symmetries, while the latter fail in the absence of distinctive local features.
To overcome these limitations, we propose a novel formulation of 6D pose estimation as a conditional flow matching problem in $\mathbb{R}^3$.
We introduce \acronym, a generative method that infers object poses via a denoising process conditioned on local features.
While prior approaches based on conditional flow matching perform denoising solely based on geometric guidance, \acronym integrates appearance-based semantic features to mitigate ambiguities caused by object symmetries.
We further incorporate RANSAC-based registration to handle outliers.
We validate \acronym on five datasets from the established BOP benchmark. \acronym outperforms prior methods with an average improvement of +4.5 Average Recall.
Project Website : https://tev-fbk.github.io/Flose/ 
\end{abstract}

\vspace{-1.5mm}
\section{Introduction}\label{sec:intro}
\vspace{-1.5mm}
While humans interact seamlessly with objects in their environments, this ability remains a fundamental challenge for robotic agents.
Object 6D pose estimation aims to bridge this gap by determining an object's position and orientation in 3D space from sensory images.
This is fundamental for robotic manipulation~\cite{tyree2022manipulation} and augmented reality~\cite{marchand2016ar}.

In recent years, the field has transitioned from hand-crafted to learning-based methods, which now achieve state-of-the-art performance~\cite{liu2025gdrnpp, wang2025hcceposebf}.
In this work, we focus on learning-based methods that operate under the \emph{model-based} \emph{instance-level} setting, and use RGBD data~\cite{doumanoglou2016recovering, konig2020hybrid, lipson2022cir, liu2025gdrnpp}.
Model-based methods assume access to a 3D representation of the object, provided as either a synthetic CAD model~\cite{tless} or a reconstructed 3D mesh~\cite{lmo, tudl, icbin, ycbv}.
Instance-level methods rely on direct supervision over a closed set of predefined object instances.
These methods can be roughly categorized as \emph{direct} or \emph{indirect}.
The former~\cite{liu2025gdrnpp, wang2021gdrn, uni6d, lipson2022cir} use neural networks to regress the object pose in an end-to-end fashion and are typically trained by minimizing a loss over the $\mathrm{SE}(3)$ manifold.
They may underperform in the case of object symmetries and suffer from lower accuracy due to the lack of explicit pixel-to-3D alignment.
The latter~\cite{wang2025hcceposebf, hu2022pfa, park2019pix2pose} extract local features to establish object-image correspondences and infer the object pose via robust registration.
These methods may fail to establish reliable correspondences in the absence of distinctive local features~\cite{hamza2025dgedi}.
Recently, generative modeling has emerged as a promising alternative for 3D point cloud registration, reformulating registration as a progressive denoising process instead of explicit correspondence estimation~\cite{sun2025rpf, li2025garf, pan2025rap}.

In this work, we formulate 6D pose estimation as a Conditional Flow Matching (CFM)~\cite{lipman2023flow} problem in $\mathbb{R}^3$~\cite{sun2025rpf} and present \acronym (\underline{Flo}w matching for 6D po\underline{se} estimation) that operates in the model-based instance-level setting.
\acronym estimates the object pose via a denoising process that learns the displacement field required to register noise samples with the object's 3D model.
Prior CFM-based methods suffer from two main limitations.
As they condition the denoising process solely on geometric guidance, it may be challenging to resolve 6D pose ambiguities in symmetric shapes where texture provides the only disambiguating cue~\cite{sun2025rpf}.
Then, as they typically recover the object pose via global alignment (e.g., SVD \cite{svd}) between input sensory data and the denoised points, they can be sensitive to displacement outliers~\cite{pan2025rap,sun2025rpf}.
We address these issues by explicitly capturing object appearance and resolving pose ambiguities arising from symmetries by injecting semantic features through a vision foundation model~\cite{dinov2}.
Then, we improve robustness to noisy correspondences and displacement outliers by replacing global alignment with a random sampling approach~\cite{ransac}, which identifies a subset of points geometrically consistent with a rigid transformation, performing registration exclusively on them.
We evaluate \acronym on five datasets from the BOP Benchmark~\cite{lmo, tless, tudl, icbin, ycbv}, spanning diverse objects and real-world conditions.
We compare \acronym against both direct~\cite{liu2025gdrnpp, lipson2022cir} and indirect~\cite{wang2025hcceposebf, hu2022pfa, haugaard2022surfemb, konig2020hybrid, park2019pix2pose, zebrapose} methods.
Compared to the leading method~\cite{hu2022pfa} training a model per dataset, \acronym achieves +4.5 Average Recall (AR).
Against the leading method~\cite{liu2025gdrnpp} training a model per object, \acronym achieves +1.2 AR while requiring reduced training and inference costs.
%
\noindent In summary, our contributions are:
\begin{itemize}[noitemsep,nolistsep,leftmargin=*]
    \item We propose \acronym, the first conditional flow matching formulation for instance-level object 6D pose estimation.
    \item We integrate semantic features from vision foundation models to disambiguate object symmetries and occlusions.
    \item We employ robust RANSAC-based registration to effectively filter outliers resulting from the denoising process.
\end{itemize}

\vspace{-1.5mm}
\section{Related work}\label{sec:related}
\vspace{-1.5mm}

\noindent\textbf{Instance-level methods for 6D pose estimation}
regress object poses from a closed set of predefined instances and can be categorized into direct and indirect approaches.
Indirect methods first establish image-object correspondences and then recover the pose using Perspective-n-Point (PnP) or RANSAC-based algorithms.
SurfEmb~\cite{haugaard2022surfemb} establishes dense image-object correspondences via surface embeddings.
ZebraPose~\cite{zebrapose} introduces a hierarchical encoding of the visible object surface, while HccePose(BF)~\cite{wang2025hcceposebf} extends this encoding to include the back surfaces in order to obtain denser correspondences.
Other indirect approaches focus on refining coarse pose predictions by enforcing consistency across objects or views~\cite{labbe2020cosypose, lipson2022cir, hu2022pfa}.
These methods remain constrained by the correspondence quality.
In contrast, direct methods predict an object's 6D pose in an end-to-end manner by minimizing a loss over the $\mathrm{SE}(3)$ manifold.
GDR-Net~\cite{wang2021gdrn} regresses poses using a differentiable patch-based PnP, 
while follow-up work, GDRNPP~\cite{liu2025gdrnpp}, achieves state-of-the-art performance by incorporating stronger backbones and domain randomization. 
These methods suffer from supervision ambiguity when handling symmetric objects that possess multiple valid poses but are typically assigned only a single ground-truth pose during training.
By formulating \acronym as conditional flow matching in $\mathbb{R}^3$, we mitigate sensitivity to local correspondence outliers through dense displacement regression and resolve the rotational ambiguities common in $\mathrm{SE}(3)$-based losses.

\noindent\textbf{Generative methods for 6D pose estimation}
use diffusion~\cite{ho2020denoising} or flow matching~\cite{lipman2023flow} to model a distribution over valid 6D poses rather than regressing a single prediction~\cite{ikeda2024diffusionnocs, zhang2024genpose, jin2025se3poseflow}.
Their primary focus is modeling pose ambiguity caused by occlusions (e.g., a mug with an occluded handle) or shape symmetries (e.g., a cereal box, a tuna can) by generating multiple pose hypotheses via stochastic sampling.
DiffusionNOCS~\cite{ikeda2024diffusionnocs} learns to denoise NOCS~\cite{wang2019nocs} coordinates to recover 6D poses.
GenPose~\cite{zhang2024genpose} frames pose estimation as a conditional diffusion process, while SE(3)-PoseFlow~\cite{jin2025se3poseflow} applies flow matching on the $\mathrm{SE}(3)$ manifold.
These methods operate in the category-level setting, relying on weaker supervision signals than instance-level approaches.
Consequently, a direct quantitative comparison with \acronym is not applicable, as we target accurate pose estimation where explicit 3D models of object instances are available.

\noindent\textbf{Flow matching for 3D tasks.}
Flow matching~\cite{lipman2023flow} has proven effective for shape assembly~\cite{sun2025rpf, li2025garf} and registration~\cite{pan2025rap} by modeling alignment as generative denoising.
GARF~\cite{li2025garf} applies flow matching over the $\mathrm{SE}(3)$ manifold to generate relative poses, RPF~\cite{sun2025rpf} operates directly in $\mathbb{R}^3$ to learn point cloud denoising without rigidity constraints.
RAP~\cite{pan2025rap} builds upon RPF by incorporating rigidity constraints into the Euler integration scheme during inference.
Similarly to RPF and RAP, we employ flow matching in $\mathbb{R}^3$.
However, we address the distinct task of object 6D pose estimation.
Unlike 3D registration, which typically aligns two point cloud fragments from the same sensor, 6D pose estimation requires aligning a clean, complete CAD model with a noisy, partial real-world observation subject to clutter and occlusion.


\vspace{-1.5mm}
\section{Method}\label{sec:method}
\vspace{-1.5mm}
\begin{figure*}[t!]
    \centering
    \begin{overpic}[width=\linewidth]{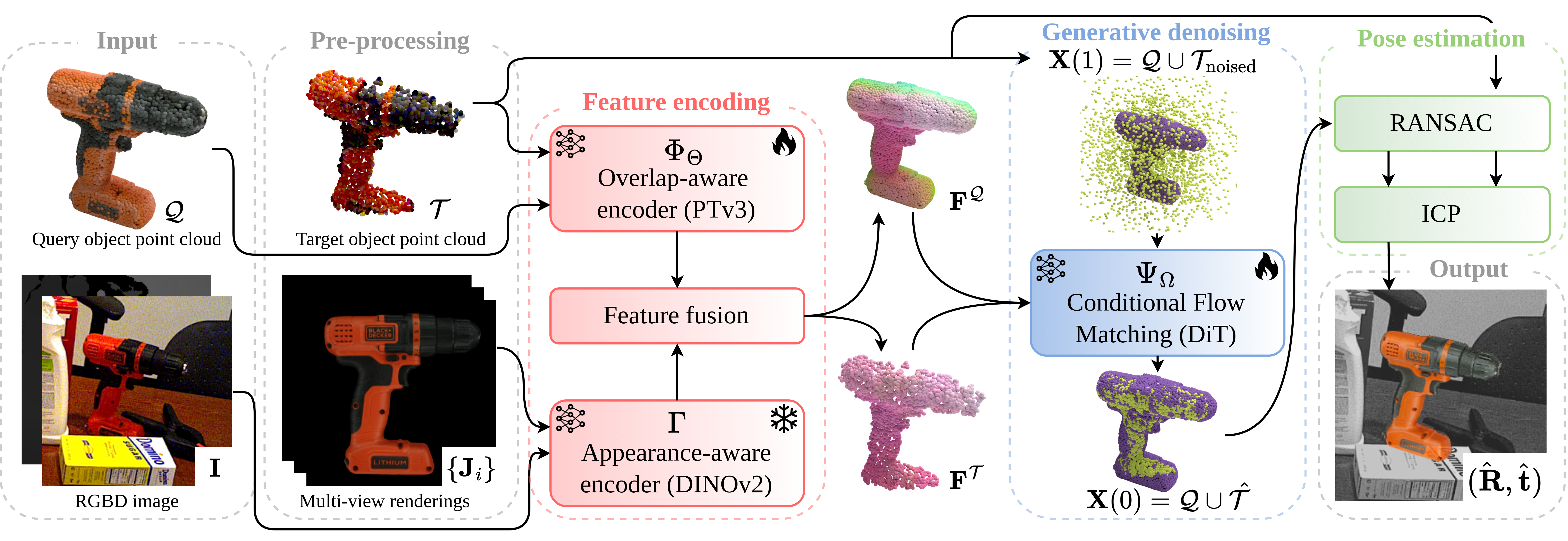}
    \end{overpic}

    \vspace{-3mm}
    \caption{
    Overview of \acronym.
    Given the query object point cloud $\mathcal{Q}$ and an RGBD image $\textbf{I}$ as input (left), \acronym estimates the 6D pose $(\hat{\textbf{R}}, \hat{\textbf{t}})$ (bottom right) through three stages: feature encoding (red), generative denoising (blue) and pose estimation (green).
    \textit{Feature encoding:} an overlap-aware encoder $\Phi_\Theta$ and appearance-aware encoder $\Gamma$ produce per-point descriptors, that are fused via a feature fusion to produce $\textbf{F}^\mathcal{Q}, \textbf{F}^\mathcal{T}$.
    Colors encode feature similarity: corresponding regions share similar colors (overlap-awareness), while semantically distinct parts differ (appearance-awareness).
    \textit{Generative denoising:} a generative network $\Psi_\Omega$, conditioned on $\textbf{F}^\mathcal{Q}, \textbf{F}^\mathcal{T}$, learns a displacement field that iteratively denoise the Gaussian noised \textbf{X}(1) into an aligned position \textbf{X}(0). 
    \textit{Pose estimation:} the 6D pose $\hat{\textbf{R}}, \hat{\textbf{t}}$ is recovered via RANSAC-based Kabsch solver followed by ICP refinement.
    }
    \label{fig:diagram}
    \vspace{-2mm}
    
\end{figure*}


\noindent\textbf{Problem formulation.}
Given the 3D model of a \emph{query object} and an RGBD image $\textbf{I}$ of a scene containing it, our goal is to recover the 6D pose of the object with respect to the camera reference frame.
We define the \emph{target object} as the partial observation of the query object visible in $\textbf{I}$.
Let $\mathcal{Q} \in \mathbb{R}^{N \times 3}$ denote the point cloud of the query object, and let $\mathcal{T} \in \mathbb{R}^{N \times 3}$ be the point cloud obtained by lifting the pixels corresponding to the target object into 3D space using the camera's intrinsic parameters.
For notational simplicity, we assume the point clouds have the same number of points $N$.
We aim to estimate the six degrees-of-freedom rigid transformation $(\hat{\mathbf{R}} \in \mathrm{SO}(3), \hat{\mathbf{t}} \in \mathbb{R}^3)$ that aligns $\mathcal{Q}$ with $\mathcal{T}$, such that $\mathcal{Q}^r = \hat{\mathbf{R}} \mathcal{Q} + \hat{\mathbf{t}} \approx \mathcal{T}$.
For the alignment process, we also define $\mathcal{T}^r$ as the target point cloud transformed into the query's canonical frame via ground truth transformation.

\noindent\textbf{Overview.}
We design a three-stage pipeline as in Fig.~\ref{fig:diagram}.
We extract point-wise features $\textbf{F}^\mathcal{Q}, \textbf{F}^\mathcal{T}$ from $\mathcal{Q}, \mathcal{T}$ by fusing overlap-aware geometric features and appearance-aware semantic features extracted with different encoders $\Phi_\Theta$ and $\Gamma$.
We deform $\mathcal{T}$ into $\hat{\mathcal{T}} \approx \mathcal{T}^r$ using a flow matching model $\Psi_\Omega$ conditioned on these features.
We estimate the rigid transformation $({\textbf{R}}, {\textbf{t}})$ that best aligns $\mathcal{T}$ with $\hat{\mathcal{T}}$ by solving a Procrustes problem.

\noindent\textbf{Feature encoding}
extracts point-level features $\textbf{F}$ by combining overlap-aware features $\textbf{O}$ and semantic features $\textbf{S}$ via point-wise addition.
The overlap-aware encoder $\Phi_\Theta$ is a parametric neural network with learnable parameters $\Theta$.
It takes 3D points (and their normals, estimated via PCA over local neighborhood) as input and predicts which points belong to the overlapping region between $\mathcal{Q}$ and $\mathcal{T}^r$.
Applying $\Phi_\Theta$ to $\mathcal{Q}$ and $\mathcal{T}$ produce binary classification logits $\textbf{L}^Q, \textbf{L}^T \in \mathbb{R}^{N \times 2}$.
During training, the parameters $\Theta$ are optimized by minimizing a binary cross-entropy loss between the predicted logits $\textbf{L}^\mathcal{Q}, \textbf{L}^\mathcal{T}$ and the ground-truth binary labels $\mathbf{y}^\mathcal{Q}, \mathbf{y}^\mathcal{T} \in [0,1]^{N}$, which indicate which points of $\mathcal{Q}$ and $\mathcal{T}$ belong to the overlap region.
$\mathbf{y}^\mathcal{Q}, \mathbf{y}^\mathcal{T}$ are constructed via nearest-neighbor search in Euclidean space: during training, $\mathcal{T}$ is first transformed to the query frame using the ground-truth transformation to obtain $\mathcal{T}^r$, then pairwise distances between $\mathcal{Q}$ and $\mathcal{T}^r$ are computed, and points with distances below a threshold $\mathcal{\epsilon}=1\%$ of the object's diameter are labeled as overlapping.
During inference, the classification head of $\Phi_\Theta$ is discarded, and only the hidden features $\textbf{O}^\mathcal{Q}, \textbf{O}^\mathcal{T} \in \mathbb{R}^{N \times F}$ are retained, yielding overlap-aware features.
However, relying solely on the geometric cues is insufficient for resolving ambiguities in objects with symmetries and sparse geometric details.
We incorporate robust appearance information by using a semantic encoder which associates pixel-level features from a frozen Vision Foundation Model (VFM) to the 3D points of $\mathcal{Q}$ and $\mathcal{T}$.
For the target object, the VFM is applied to a regular crop of $\textbf{I}$ centered on its barycenter and uses the known pixel-to-point correspondence to associate the pixel-level features with the points of $\mathcal{T}$, producing point-level features $\textbf{S}^\mathcal{T} \in \mathbb{R}^{N \times G}$, where $G$ is the dimensionality of the VFM features.
For the query object, the encoder generates multi-view renderings $\{\mathbf{J}_i\}$ of its 3D model, applies the VFM to the synthetic images, establishes pixel-to-point correspondences using the synthetic camera intrinsics, and associates the pixel-level features to the corresponding 3D points.
If multiple pixels map to the same 3D point, their features are fused via average pooling.
This process is performed once per object during pre-processing; at inference, the query features $\textbf{S}^\mathcal{Q} \in \mathbb{R}^{N \times G}$ are loaded from memory.
The dimensionality of $\textbf{S}^\mathcal{Q}$ and $\textbf{S}^\mathcal{T}$ is reduced to $F$ using PCA, to match that of the overlap-aware features.
Specifically, we compute the PCA basis only on $\textbf{S}^\mathcal{Q}$ and apply this projection to both $\textbf{S}^\mathcal{Q}$ and all instances of $\textbf{S}^\mathcal{T}$.
Finally, the output features are defined by combining overlap-aware and semantic features via point-wise addition, i.e., $\textbf{F} = \mathrm{norm}( \mathrm{norm}(\textbf{O}) + \mathrm{norm}(\textbf{S}) ) \in \mathbb{R}^{N \times F}$, where $\mathrm{norm}(\cdot)$ denotes the operator that normalizes each point feature to have unit $L_2$ norm.

\noindent\textbf{Conditional flow matching.}
We use conditional flow matching framework~\cite{sun2025rpf} to transform source point cloud $\textbf{X}(0)$ to target $\textbf{X}(1)$ via linear interpolation $\textbf{X}(t) = (1-t) \textbf{X}(0) + t \textbf{X}(1)$, $t \in [0, 1]$.
In our setting, $\textbf{X}(0) = \mathcal{Q} \cup \mathcal{T}^r \in \mathbb{R}^{2N \times 3}$, while $\textbf{X}(1) \in \mathbb{R}^{2N \times 3}$ is obtained by uniformly sampling Gaussian noise in 3D space.
The neural network $\Psi_\Omega( t, \textbf{X}(t) | \textbf{C} )$ learns to reverse the RPF process by predicting a vector field $\textbf{V} \approx \textbf{X}(0) - \textbf{X}(t) \in \mathbb{R}^{2N \times 3}$ that transforms $\textbf{X}(t)$ into $\textbf{X}(0)$.
$\Psi_\Omega$ is conditioned on overlap-aware and semantic features $\textbf{C}$.
The parameters $\Omega$ are optimized using the conditional flow matching loss~\cite{sun2025rpf}.
At inference time, we perform $K$ uniform Euler steps as $\hat{\textbf{X}}(t - \Delta t) = \hat{\textbf{X}}(t) - \Psi_\Omega(t, \textbf{X}(t) | \textbf{C} ) \Delta t$, where $\Delta t = 1 / K$.
After $K$ iterations, the resulting $\hat{\textbf{X}}(0) = \mathcal{Q} \cup \hat{\mathcal{T}}$ approximates $\textbf{X}(0) = \mathcal{Q} \cup \mathcal{T}^r$.
In practice, we replace the components of $\textbf{V}$ corresponding to the anchor $\mathcal{Q}$ with a null field and apply the resulting velocity field to $\mathcal{T}$ to obtain a deformed version $\hat{\mathcal{T}} = \mathcal{T} + \textbf{V} \approx \mathcal{T}^r$.
Unlike~\cite{sun2025rpf}, which conditions $\Psi_\Omega$ solely on geometric positional encoding, we enrich the conditioning signal $\textbf{C} \in \mathbb{R}^{N \times (F+H)}$ with overlap-aware and semantic point-wise features that jointly capture geometric structure and object appearance.
Specifically, $\textbf{C} = [ \textbf{F} \in \mathbb{R}^{N \times F} | \textbf{P} \in \mathbb{R}^{N \times H} ]$, where $\textbf{F}$ are extracted by the feature encoder, $\textbf{P}$ are obtained by the positional encoder $\Xi$, and $|$ denotes concatenation along the last dimension.
For each point, $\Xi$ encodes a 10-dimensional array composed of concatenated coordinates, normals, the coordinates of the corresponding point in the noisy initialization, and a binary scalar identifying either $\mathcal{Q}$ or $\mathcal{T}$.
In practice, $\Xi$ applies sinusoidal embeddings with logarithmically spaced frequencies to each geometric attribute independently~\cite{nerf}.
The resulting embeddings are concatenated with the learned point-wise features and a linear projection is then applied to obtain the final positional encoding \textbf{P} used to condition the flow model.

\begin{figure}[t!] 
\centering

\begin{tikzpicture} [font=\scriptsize]
    \begin{axis}[
        height=20mm,
        width=\columnwidth,
        bar width=20pt,
        ymin=17,
        ymax=89,
        ylabel={IR},
        ylabel style={font=\scriptsize},
        ytick={0.2,0.4,0.6,0.8,1.0},
        yticklabels={}, 
        xmode=log,
        log basis x={10},
        xlabel={Spatial threshold $\tau$ as \% of object diameter},
        xlabel style={font=\scriptsize},
        xlabel style={yshift=0.5em},
        xtick={0.005, 0.01, 0.02, 0.03, 0.05, 0.1},
        xticklabels={$0.5\%$, $1\%$, $2\%$, $3\%$, $5\%$, $10\%$},
        nodes near coords={
            \pgfmathprintnumber[fixed, fixed zerofill, precision=0]{\pgfplotspointmeta}\%
        },
        nodes near coords align={vertical},
        nodes near coords style={font=\scriptsize},
        axis lines=left,
        enlarge x limits=0.08,
    ]
    \addplot coordinates {
        (0.005, 18)
        (0.01, 38)
        (0.02, 65)
        (0.03, 75)
        (0.05, 82)
        (0.1, 88)
    };   
    \end{axis}
\end{tikzpicture}

\vspace{-3mm}
\caption{
Sensitivity of the Inlier Ratio (IR) to the spatial threshold $\tau$.
IR is the percentage of points in $\hat{\mathcal{T}}$ whose distance to the corresponding points in $\mathcal{T}^r$ is smaller than $\tau$.
At strict thresholds, the majority ($>80\%$) of correspondences are outliers.
}
\label{fig:inlier_ratio}
\vspace{-1.0mm}
\end{figure}
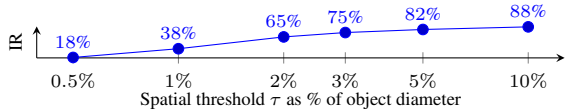

\noindent\textbf{Pose estimation.}
$\Psi_\Omega$ predicts $\hat{\mathcal{T}}$, a non-linear deformation of $\mathcal{T}$ that approximates its rigidly aligned counterpart $\mathcal{T}^r$.
Since flow fields are predicted per point without explicit rigidity constraints, the resulting  $\hat{\mathcal{T}}$ is not a perfect rigid transformation of $\mathcal{T}$.
Fig.~\ref{fig:inlier_ratio} quantifies this effect: at a strict
threshold $\tau = 0.5\%$ of the object's diameter, only $18\%$ of correspondences are inliers,
and even at $\tau = 20\%$ nearly $35\%$ of points remain outliers on all instances of LM-O.
Global alignment via SVD over all correspondences is thus dominated by outliers, yielding inaccurate poses.
We therefore employ RANSAC, where each iteration samples a minimal set of three correspondences to solve the orthogonal Procrustes problem via the Kabsch algorithm.
The candidate transformation with most inliers is selected as the initial estimate $({\textbf{R}},{\textbf{t}})$, representing the transformation from target to query frame.
We refine this pose using ICP.
Finally, we invert this transformation to obtain the standard 6D pose of the query object in the camera frame. 

\vspace{-1.5mm}
\section{Experiments}\label{sec:exps}
\vspace{-1.5mm}

\noindent\textbf{Implementation details.}
During pre-processing, the target object in the input image is segmented using the ZebraPose~\cite{zebrapose} masks publicly available on the BOP leaderboard.
In the feature extraction stage, $\Phi_\Theta$ is based on PointTransformerV3~\cite{ptv3} and produces overlap-aware features of size $F=64$, while
$\Gamma$ is based on a frozen DINOv2-Giant~\cite{dinov2}, yielding appearance-aware features of dimension $G = 1536$, which are reduced to $F$ via PCA.
Finally, these features are fused via point-wise summation.
In the flow matching stage, overlap- and semantic features are concatenated with position-aware features of size $H=210$, derived from the positional encoding of object coordinates and normals, and noisy point positions.
The resulting features of size $F+H=274$ are fed to $\Psi_\Omega$, which is based on Diffusion Transformers~\cite{dit}.
During training, $\Theta$ and $\Omega$ are initialized with RPF~\cite{sun2025rpf} checkpoints and finetuned on the training split of each dataset in two separate phases for 200 and 100 epochs, respectively.
Training is distributed across 12 NVIDIA A100 GPUs with an effective batch size of 384.
During post-processing, we execute RANSAC-based registration for 1k iterations with an inlier threshold of 1cm, followed by ICP-based refinement for 3k iterations with a distance threshold of 1cm.

\smallskip
\noindent\textbf{Experimental validation.}
We validate \acronym on five datasets from the BOP Benchmark~\cite{lmo, tless, tudl, icbin, ycbv}, featuring diverse object types (everyday vs.~industrial items), varying appearance (textured vs.~texture-less), and complex symmetries, as well as environmental challenges such as heavy occlusion and changing lighting conditions.
We compare \acronym against ten peer-reviewed competitors, categorized into two protocols:
(i) \emph{per-object} methods that train a dedicated model for each object instance, such as ZebraPose~\cite{zebrapose}, GDRNPP~\cite{liu2025gdrnpp}, and HccePose(BF)~\cite{wang2025hcceposebf};
and (ii) \emph{per-dataset} methods that train a single model for all objects within a dataset, such as SurfEmb~\cite{haugaard2022surfemb}, CIR~\cite{lipson2022cir}, and PFA~\cite{hu2022pfa}.
We adhere to the latter setting.
Following the standard BOP protocol, we measure pose accuracy using the Average Recall (AR) metric~\cite{hodan2024bop}.

\begin{table}[t]
\tabcolsep 4pt
\caption{
Comparison with RGBD methods in terms of AR.
The top block (rows 1-5) trains a separate model for each object, while the bottom block (rows 6-11) trains a Single Model (S.M.) per dataset.
Row 12 quantifies the absolute improvement of \acronym over the strongest S.M. competitor.
}
\vspace{1mm}
\label{tab:quant}
\resizebox{\columnwidth}{!}{%
\begin{tabular}{rlccccccc}
    \toprule
    & Method & S.M. & LM-O & T-LESS & TUD-L & IC-BIN & YCB-V & Avg \\
    \toprule
    {\color{gray}\scriptsize 1} & Pix2Pose~\cite{park2019pix2pose} & & 58.8 & 51.2 & 82.0 & 39.0 & 78.8 & 62.0 \\
    {\color{gray}\scriptsize 2} & ZebraPose~\cite{zebrapose} & & 75.2 & 72.7 & 94.8 & 65.2 & 86.6 & 78.9 \\
    {\color{gray}\scriptsize 3} & GDRNPP (BOP22)~\cite{liu2025gdrnpp} & & 77.5 & 87.4 & 96.6 & 72.2 & 92.1 & 85.2 \\
    {\color{gray}\scriptsize 4} & HccePose(BF)~\cite{wang2025hcceposebf} & & 80.5 & 87.9 & 94.4 & 72.4 & 91.1 & 85.3 \\
    {\color{gray}\scriptsize 5} & GDRNPP (BOP23)~\cite{liu2025gdrnpp} & & 79.4 & 91.4 & 96.4 & 73.7 & 92.8 & 86.7 \\
    \midrule
    {\color{gray}\scriptsize 6} & Koenig-Hybrid~\cite{konig2020hybrid} & \cmark & 63.1 & 65.5 & 92.0 & 43.0 & 70.1 & 66.7 \\
    {\color{gray}\scriptsize 7} & CosyPose~\cite{labbe2020cosypose} & \cmark & 71.4 & 70.1 & 93.9 & 64.7 & 86.1 & 77.2 \\
    {\color{gray}\scriptsize 8} & SurfEmb~\cite{haugaard2022surfemb} & \cmark & 75.8 & 83.3 & 93.3 & 65.6 & 82.4 & 80.1 \\
    {\color{gray}\scriptsize 9} & CIR~\cite{lipson2022cir} & \cmark & 73.4 & 77.6 & 96.8 & 67.6 & 89.3 & 81.0 \\
    {\color{gray}\scriptsize 10} & PFA~\cite{hu2022pfa} & \cmark & 79.7 & 85.0 & 96.0 & 67.6 & 88.8 & 83.4 \\
    \rowcolor{tableazure} {\color{gray}\scriptsize 11} & \acronym (ours) & \cmark & \textbf{86.1} & \textbf{86.9} & \textbf{98.8} & \textbf{74.8} & \textbf{92.8} & \textbf{87.9} \\
    \rowcolor{tableazure} {\color{gray}\scriptsize 12} & Improv. over row 10 & & \color{forestgreen} +6.4 &\color{forestgreen} +1.9 &\color{forestgreen} +2.8 &\color{forestgreen} +7.2 &\color{forestgreen} +4.0 & \color{forestgreen} +4.5 \\
    \bottomrule
\end{tabular}
\vspace{-4mm}
}
\end{table}


\smallskip
\noindent\textbf{Quantitative results.}
Tab.~\ref{tab:quant} details pose accuracy (AR) for various instance-level RGBD methods.
Competitor results are reported from~\cite{liu2025gdrnpp, wang2025hcceposebf}.
The bottom block (rows 6-11) lists per-dataset methods that are directly comparable with \acronym, as they train a single model per dataset (\cmark{} in S.M. column, 5 models total).
\acronym achieves an average improvement of +4.5 AR over PFA~\cite{hu2022pfa}, the strongest competitor in this category.
The performance gain is consistent across all datasets (row 12).
The top block (rows 1-5) shows per-object methods that train one model per object (54 models total).
Even in this more challenging comparison, \acronym outperforms the current state-of-the-art GDRNPP~\cite{liu2025gdrnpp} by +1.2 AR on average.
We noted that the average gain in AR is higher for symmetric objects i.e. +3.95 AR for LM-O (Eggbox and Glue bottle) against +2.12 AR on other objects, highlighting the role of semantic features for solving ambiguities.
\acronym achieves this while relying on less specialized models (the same model is used for all object instances of a dataset) and requiring less compute and memory resources (54/5~$\approx$~11$\times$ fewer models to train and load).

\begin{figure}[t!]
    \centering

    \begin{minipage}{0.1\columnwidth}
    \end{minipage}%
    \begin{minipage}{0.3\columnwidth}
    \footnotesize \hspace{5mm} Input image
    \end{minipage}%
    \begin{minipage}{0.3\columnwidth}
    \footnotesize \hspace{6mm} \acronym (ours)
    \end{minipage}%
    \begin{minipage}{0.3\columnwidth}
    \footnotesize \hspace{10mm} RPF~\cite{sun2025rpf}
    \end{minipage}

    \begin{minipage}{0.035\columnwidth}
    \rotatebox{90}{\footnotesize Ape (LM-O)}
    \end{minipage}%
    \begin{minipage}{0.965\columnwidth}
    \centering
    \includegraphics[width=\columnwidth]{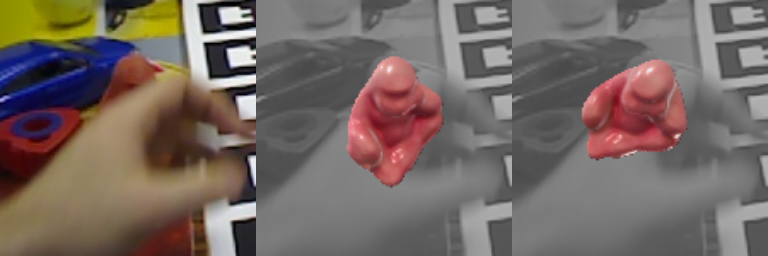}
    \end{minipage}
    
    \begin{minipage}{0.035\columnwidth}
    \rotatebox{90}{\footnotesize Watering can (LM-O)}
    \end{minipage}%
    \begin{minipage}{0.965\columnwidth}
    \centering
    \includegraphics[width=\columnwidth]{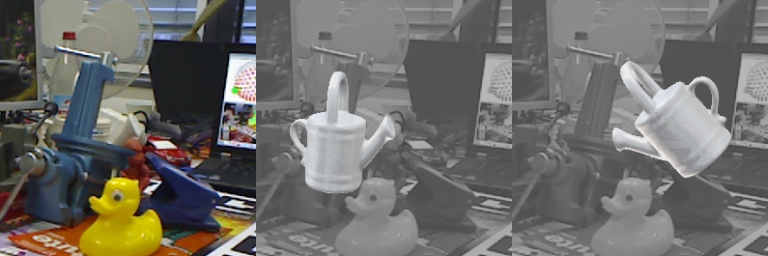}
    \end{minipage}


    \begin{minipage}{0.035\columnwidth}
    \rotatebox{90}{\footnotesize Mug (YCB-V)}
    \end{minipage}%
    \begin{minipage}{0.965\columnwidth}
    \centering
    \includegraphics[width=\columnwidth]{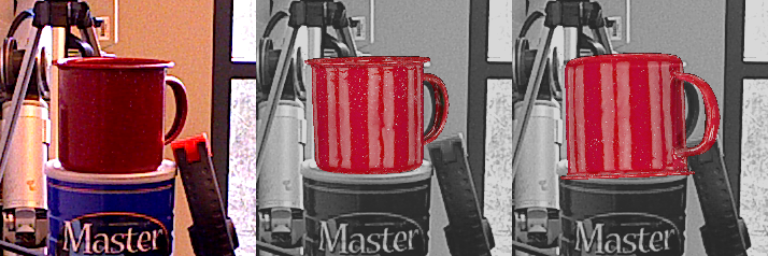}
    \end{minipage}

    \begin{minipage}{0.035\columnwidth}
    \rotatebox{90}{\footnotesize Can (YCB-V)}
    \end{minipage}%
    \begin{minipage}{0.965\columnwidth}
    \centering
    \includegraphics[width=\columnwidth]{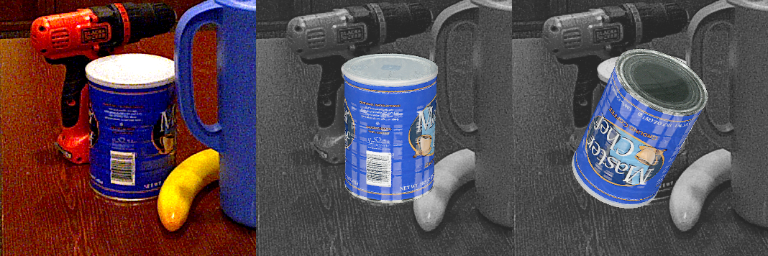}
    \end{minipage}

    \vspace{-0.5mm}
    \caption{
    Qualitative comparison of \acronym (center) vs. an RPF-based~\cite{sun2025rpf} baseline adapted for pose estimation (right).
    By integrating semantic features and outlier-robust registration, \acronym predicts more accurate poses under severe occlusions (rows 1-2) and resolves symmetry ambiguities where pure geometric methods fail (rows 3-4).
    }
    \label{fig:qual}
    \vspace{-4.6mm}
\end{figure}

\smallskip
\noindent\textbf{Qualitative results.}
Fig.~\ref{fig:qual} presents qualitative comparisons on LM-O and YCB-V images (first column) between \acronym (second column) and an RPF-based~\cite{sun2025rpf} baseline finetuned on the same data (third column).
Each image shows a rendering of the object's 3D model in the predicted pose overlaid with a grayscale version of the input image; prediction accuracy is assessed by measuring the alignment between the colored rendering and the visible portion of the grayscale object in the background image.
Rows 1-2 show a red toy ape and a white watering can under severe occlusions caused by hand-object interaction and clutter.
Rows 3-5 show objects with varying degrees of symmetry:
a white glue bottle with a symmetric shape but asymmetric texture (front vs.~back stickers),
a red mug with uniform texture,
and a blue Master Chef can with a cylindrical shape (infinite rotational symmetry) and rich texture (brand name and logo).
\acronym outperforms RPF across all scenarios, showing its robustness to real-world noise.

\begin{figure}[t!]
    \centering

    %
    %
    \scriptsize (a) Impact of different types of feature conditioning

    \begin{tikzpicture}[font=\scriptsize]
    \begin{axis}[
        ybar,
        height=20mm,
        width=\columnwidth,
        bar width=20pt,
        ymin=70.1,
        ymax=87.1,
        ylabel={AR},
        symbolic x coords={Appearance only, Overlap only, Both (ours)},
        xtick=data,
        ytick=\empty,
        yticklabels={},
        nodes near coords,
        nodes near coords align={vertical},
        nodes near coords style={font=\scriptsize},
        axis lines=left,
        enlarge x limits=0.25,
    ]
    \addplot coordinates {
        (Appearance only, 71.1)
        (Overlap only, 83.5)
        (Both (ours), 86.1)
    };
    \end{axis}
    \end{tikzpicture}

    \vspace{-1mm}
    \begin{tikzpicture}[font=\scriptsize]
    \begin{axis}[
        height=20mm,
        width=\columnwidth,
        bar width=20pt,
        ymin=6.2,
        ymax=31.0,
        ylabel={IR gain},
        xlabel={Spatial threshold $\tau$ as \% of object diameter},
        xlabel style={yshift=0.5em},
        symbolic x coords={0.005, 0.01, 0.02, 0.03, 0.05, 0.1},
        xtick=data,
        xticklabels={0.5\%, 1\%, 2\%, 3\%, 5\%, 10\%},
        ytick=\empty,
        yticklabels={},
        nodes near coords={
            $+\pgfmathprintnumber{\pgfplotspointmeta}\%$
        },
        nodes near coords align={vertical},
        nodes near coords style={font=\scriptsize},
        axis lines=left,
        enlarge x limits=0.08,
    ]
    \addplot coordinates {
        (0.005, 30)
        (0.01, 18)
        (0.02, 14)
        (0.03, 12)
        (0.05, 10)
        (0.1, 7)
    };
    \end{axis}
    \end{tikzpicture}

    %
    %
    \smallskip
    \scriptsize (b) Comparison of pose estimation and refinement strategies

    \begin{tikzpicture}[font=\scriptsize]
    \begin{axis}[
        ybar,
        height=20mm,
        width=\columnwidth,
        bar width=20pt,
        ymin=80,
        ymax=87.1,
        ylabel={AR},
        symbolic x coords={SVD, SVD + ICP, RANSAC, RANSAC + ICP},
        xtick=data,
        xlabel={},
        ytick=\empty,
        nodes near coords={
            \pgfmathprintnumber[fixed, fixed zerofill, precision=1]{\pgfplotspointmeta}
        },
        nodes near coords align={vertical},
        nodes near coords style={font=\scriptsize},
        axis lines=left,
        enlarge x limits=0.2,
    ]
    \addplot coordinates {
        (SVD, 81.0)
        (SVD + ICP, 84.2)
        (RANSAC, 81.8)
        (RANSAC + ICP, 86.1)
    };
    \end{axis}
    \end{tikzpicture}

    %
    %
    \smallskip
    \scriptsize (c) Effect of denoising iterations on AR and inference time

    \begin{tikzpicture}[font=\scriptsize]
    \begin{axis}[
        height=20mm,
        width=\columnwidth,
        bar width=20pt,
        ymin=72.6,
        ymax=85,
        ylabel={AR},
        xlabel style={yshift=0.5em},
        symbolic x coords={1, 10, 20, 30, 40, 50, 70, 100},
        xtick=data,
        ytick=\empty,
        yticklabels={},
        nodes near coords={
            \pgfmathprintnumber[fixed, fixed zerofill, precision=1]{\pgfplotspointmeta}
        },
        nodes near coords align={vertical},
        nodes near coords style={font=\scriptsize},
        axis lines=left,
        enlarge x limits=0.08,
    ]
    \addplot coordinates {
        (1, 73.6)
        (10, 79.4)
        (20, 80.2)
        (30, 81.3)
        (40, 82.4)
        (50, 83.5)
        (70, 83.7)
        (100, 84.0) 
    };
    \end{axis}
    \end{tikzpicture}

    \vspace{-1mm}
    \begin{tikzpicture}[font=\scriptsize]
    \begin{axis}[
        height=20mm,
        width=\columnwidth,
        bar width=20pt,
        ymin=0.08,
        ymax=1.73,
        ylabel={Seconds},
        xlabel={Euler integration steps},
        xlabel style={yshift=0.5em},
        symbolic x coords={1, 10, 20, 30, 40, 50, 70, 100},
        xtick=data,
        ytick=\empty,
        yticklabels={},
        nodes near coords={
            \pgfmathprintnumber[fixed, fixed zerofill, precision=1]{\pgfplotspointmeta}
        },
        nodes near coords align={vertical},
        nodes near coords style={font=\scriptsize},
        axis lines=left,
        enlarge x limits=0.08,
    ]
    \addplot coordinates {
        (1, 0.09)
        (10, 0.17)
        (20, 0.34)
        (30, 0.52)
        (40, 0.69)
        (50, 0.86)
        (70, 1.20)
        (100, 1.72) 
    };
    \end{axis}
    \end{tikzpicture}

    \vspace{-3mm}
    \caption{
    Ablation study on LM-O:
    (a) Impact of the conditioning features on the flow matching process, measured in terms of AR (top) and IR gain relative to a baseline using only overlap-aware features (bottom);
    (b) Comparison of pose solvers (SVD vs.~RANSAC) and the effect of ICP-based refinement.
    Hyperparameter study on LM-O: 
    (c) Pose accuracy and inference time as a function of Euler integration steps.
    }
    \label{fig:ablation}
    \vspace{-4.0mm}
\end{figure}
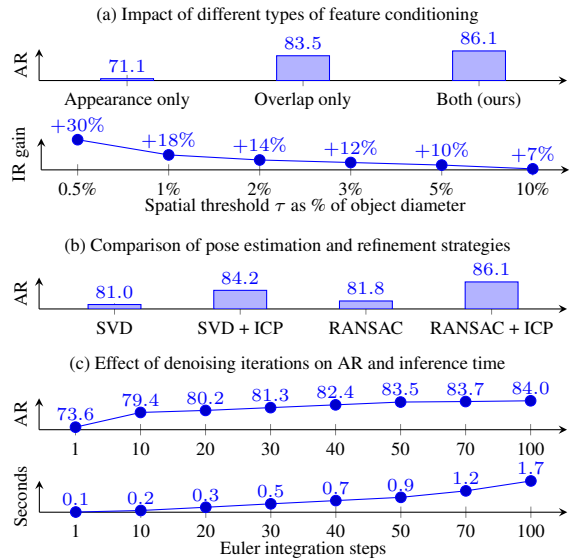


\smallskip
\noindent\textbf{Ablation and hyperparameter study}
on LM-O using ZebraPose~\cite{zebrapose} segmentation masks.
Fig.~4(a) analyzes the impact of feature conditioning.
The top plot shows that fusing appearance-based~\cite{dinov2} and overlap-aware~\cite{sun2025rpf} features surpasses either modality in isolation, yielding gains of +15.0 and +2.6 AR, respectively.
The bottom plot reveals the source of this improvement: the fused features consistently boost the Inlier Ratio (IR) relative to the geometry-only baseline.
The gain is most pronounced at the strictest thresholds, confirming that semantic information provides the discriminative cues necessary for establishing accurate correspondence.
Fig.~4(b) compares different pose estimation strategies, showing that RANSAC outperforms SVD by robustly filtering outliers caused by denoising outliers. 
Furthermore, applying ICP-based geometric refinement boosts accuracy by an additional +4.3 AR, effectively correcting residual alignment errors from the coarse pose.
Fig.~4(c) analyzes the impact of the Euler integration steps.
The AR improves with the number of iterations, saturating at higher counts.
As shown in the bottom plot, the average inference time per image scales linearly from 0.1 to 1.7 seconds.
This hyper-parameter allows balancing accuracy against computational efficiency.

\vspace{-1.5mm}
\section{Conclusions}\label{sec:conclusion}
\vspace{-1.5mm}

We presented \acronym, the first instance-level 6D pose estimation method formulated as conditional flow matching in $\mathbb{R}^3$.
\acronym conditions the generative denoising process on a fusion of overlap-aware and semantic features, and employs robust RANSAC-based registration.
Experiments demonstrate that \acronym outperforms state-of-the-art competitors, 
achieving superior robustness to symmetries and occlusions while requiring less training and smaller inference costs.
Our iterative formulation enables explicit control over the accuracy–efficiency trade-off by adjusting the number of Euler integration steps.

\noindent\textbf{Limitations and future work.}
(1) Our current pipeline relies on a two-stage training process, which could be simplified using off-the-shelf geometric and semantic descriptors. 
(2) The iterative nature of flow matching hinders time-critical applications; single-step denoising would enable these applications.
(3) \acronym depends on object-level segmentation performance; extending it to operate at scene-level is another interesting research direction.




\bibliographystyle{main/IEEEbib}
\bibliography{main.bib}

\end{document}